\documentclass{bytedance}
\usepackage[toc,page,header]{appendix}

\usepackage{minitoc}
\usepackage{amsfonts}
\usepackage{amssymb}
\usepackage{tabularx}
\usepackage{cancel}

\usepackage{tabulary,multirow,xspace}
\usepackage{fixmath,mathtools,nicefrac,mmstyle}
\usepackage{subcaption}
\usepackage{caption}
\usepackage[misc]{ifsym} 
\usepackage{colortbl}
\usepackage{algorithm}
\usepackage{algorithmicx}
\usepackage{algpseudocode}
\usepackage{wrapfig}
\usepackage{multicol}
\usepackage[most]{tcolorbox}
\usepackage{pifont}
\usepackage{tikz}
\usepackage{adjustbox}
\definecolor{codegreen}{rgb}{0,0.6,0}
\definecolor{codegray}{rgb}{0.5,0.5,0.5}
\definecolor{codepurple}{rgb}{0.58,0,0.82}
\definecolor{backcolour}{rgb}{0.95,0.95,0.92}
\definecolor{boxblue}{RGB}{57,89,163}
\definecolor{boxbluebg}{RGB}{230,237,250}

\lstdefinestyle{mystyle}{
    backgroundcolor=\color{backcolour},   
    commentstyle=\color{codegreen},
    keywordstyle=\color{magenta},
    numberstyle=\tiny\color{codegray},
    stringstyle=\color{codepurple},
    basicstyle=\ttfamily\footnotesize,
    breakatwhitespace=false,         
    breaklines=true,                 
    captionpos=b,                    
    keepspaces=true,                 
    numbers=none,                    
    numbersep=5pt,                  
    showspaces=false,                
    showstringspaces=false,
    showtabs=false,                  
    tabsize=2
}
\usepackage{pifont}

\newlength\savewidth
\newcolumntype{x}[1]{>{\centering\arraybackslash}p{#1pt}}

\newcommand{\app}{\raise.17ex\hbox{$\scriptstyle\sim$}}

\makeatletter
\DeclareRobustCommand\onedot{\futurelet\@let@token\@onedot}
\def\@onedot{\ifx\@let@token.\else.\null\fi\xspace}

\makeatother

\makeatletter

\newcommand{\Rmnum}[1]{\expandafter\@slowromancap\romannumeral #1@}
\makeatother

\usepackage{xcolor}
\usepackage{graphicx}
\usepackage{amssymb}
\usepackage{pifont}
\usepackage{amsmath} 
\usepackage{float}
\usepackage{wrapfig}
\usepackage{multirow}
\usepackage{tcolorbox}
\tcbuselibrary{breakable, skins, raster}
\usepackage{listings}
\usepackage{listings}

\definecolor{commentgreen}{rgb}{0.1, 0.4, 0.1}
\definecolor{keywordblue}{rgb}{0.1, 0.1, 0.7}
\definecolor{stringred}{rgb}{0.7, 0.1, 0.1}

\lstdefinestyle{mystyle}{
    commentstyle=\color{commentgreen},
    keywordstyle=\color{keywordblue},   
    stringstyle=\color{stringred},
    basicstyle=\ttfamily\scriptsize, 
    breaklines=true,
    keepspaces=true,
    showstringspaces=false,
    frame=none,                     
    language=Python, 
}

\newcommand{\name}{MoGA}
\title{\name{}: Mixture-of-Groups Attention for End-to-End Long Video Generation}

\author[1,*]{Weinan Jia}
\author[2,\dagger]{Yuning Lu}
\author[1]{Mengqi Huang}
\author[3,*]{Hualiang Wang}
\author[4,*]{Binyuan Huang} 
\author[1]{\mbox{Nan Chen}}
\author[2]{Mu Liu}
\author[2]{Jidong Jiang}
\author[1,\dagger]{Zhendong Mao}

\affiliation[1]{University of Science and Technology of China}
\affiliation[2]{FanqieAI, ByteDance China}
\affiliation[3]{Hong Kong University of Science and Technology}
\affiliation[4]{Wuhan University}

\contribution[*]{Work done at FanqieAI, ByteDance China}
\contribution[\dagger]{Corresponding authors}

\abstract{
Long video generation with Diffusion Transformers (DiTs) is bottlenecked by the quadratic scaling of full attention with sequence length. Since attention is highly redundant, outputs are dominated by a small subset of query–key pairs. Existing sparse methods rely on blockwise coarse estimation, whose accuracy–efficiency trade-offs are constrained by block size. This paper introduces Mixture-of-Groups Attention (\textbf{MoGA}), an efficient sparse attention that uses a lightweight, learnable token router to precisely match tokens without blockwise estimation. Through semantic-aware routing, MoGA enables effective long-range interactions. As a kernel-free method, MoGA integrates seamlessly with modern attention stacks, including FlashAttention and sequence parallelism. Building on MoGA, we develop an efficient long video generation model that end-to-end produces minute-level, multi-shot, 480p videos at 24 fps, with a context length of approximately 580k. Comprehensive experiments on various video generation tasks validate the effectiveness of our approach.
}

\date{\today}
\project{\href{https://jiawn-creator.github.io/mixture-of-groups-attention/}{\textbf{MoGA}}}

\begin{document}
\maketitle

\section{Introduction}

A growing body of research indicates that scaling laws are a primary driver of progress toward artificial general intelligence~\citep{brown2020language,team2023gemini,kaplan2020scaling}.
As model parameters and data scale to billions, Transformer-based foundation models~\citep{vaswani2017attention} often exhibit emergent capabilities~\citep{wei2022emergent, kaplan2020scaling,radford2021learning}.
In video generation, given the inherently temporal nature, progress requires not only scaling parameters and data but, more critically, scaling the effective context length.
This need is especially salient for long-form video generation (e.g.,~movies), where persistent memory is essential for maintaining consistency of environments and characters~\citep{yu2025context}.

The main challenge of vanilla attention~\citep{vaswani2017attention} for long sequences is its computational cost, which grows quadratically with the context length.
To mitigate the challenge, prior work~\citep{zhuang2024vlogger, tian2024videotetris, huang2025filmaster, xiao2025captain,wang2025storyanchors} adopts a multi-stage pipeline that first generates key frames and then synthesizes intermediate frames. 
However, this design yields disjoint objectives that are not directly optimized for the end task, leading to error accumulation across stages.
It also introduces hand-crafted inductive biases, hindering scalability.

\begin{figure*}[t]
  \centering
  \includegraphics[width=\linewidth]{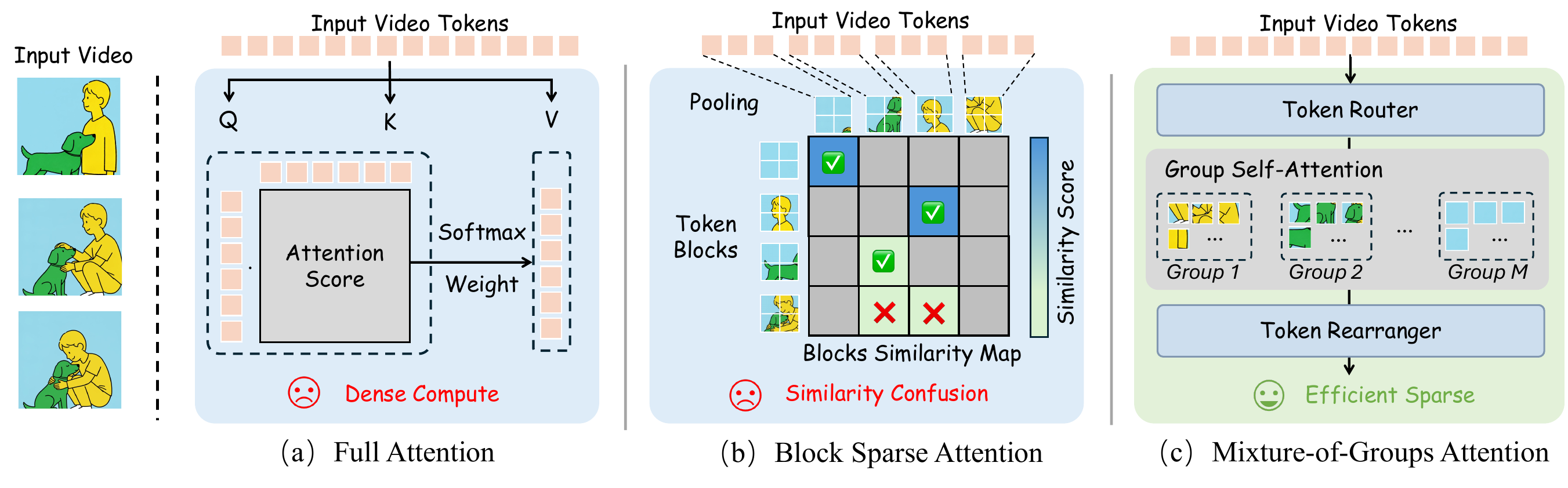}
  \vspace{-15pt}
  \caption{Illustration of our motivation.
  (a) Full attention suffers from dense computing when dealing with long sequences. 
  (b) Block sparse attention~\citep{lu2025moba} may fail when block-level similarity is confused, resulting in unreliable attention. (c) Mixture-of-Groups attention uses a lightweight token router (i.e., a \textit{single linear layer}) that assigns tokens to specialized groups, enabling groupwise attention and efficient long-context modeling.
}
  \label{fig:motivation}
  \vspace{-10pt}
\end{figure*}

For \textit{end-to-end} long video generation, one line of work compresses historical content to accommodate longer contexts (e.g., via recurrent layers~\citep{dalal2025one} or FramePack~\citep{zhang2025packing}), which inevitably results in information loss.
A complementary direction exploits \textit{sparse attention}~\citep{zaheer2020big} by restricting computation to a selected subset of salient query–key pairs. 
Existing selection strategies generally fall into two categories: (i) \textit{static} selection, i.e., prior-driven heuristics that emphasize local spatiotemporal neighborhoods, which is efficient but limited in capturing dynamic long-range dependencies~\citep{li2025radial, xi2025sparse, gao2025seedance, seawead2025seaweed}; and (ii) coarse-to-fine \textit{dynamic} selection, which first estimates block-level importance scores, routes query tokens to the top-k blocks, and then applies fine-grained attention within the selected blocks~\citep{wu2025vmoba, cai2025mixture, yang2025sparse,yuan2025native,lu2025moba}. 
As shown in Fig.\ref{fig:motivation}(b), the latter introduces an efficiency-performance trade-off: using larger blocks with a small top-k reduces the computational cost of the coarse stage but reduces selection performance.

In this work, we reveal that such coarse-grained estimation is unnecessary and each token should be precisely allocated. To achieve this, we propose Mixture-of-Groups Attention (\textbf{MoGA}), a simple and efficient dynamic token routing solution for end-to-end long video generation. A lightweight router (i.e., a \textit{single linear layer}) is employed to assign tokens to specific groups, as illustrated in Fig. \ref{fig:motivation}(c), inspired by the Mixture-of-Experts (MoE)~\citep{jacobs1991adaptive}. Full attention is then performed within each group, where the groupwise attention integrates seamlessly with modern attention kernels, e.g.,~FlashAttention~\citep{dao2023flashattention}.
Intuitively, the linear router's weights can be viewed as implicit cluster centers, enabling direct assignment of tokens to learnable anchors, without global similarity estimation.
Furthermore, to balance long-range coherence and local fidelity, we couple MoGA with the spatiotemporal window attention~\citep{gao2025seedance}, which can be considered as groupwise attention with static, predefined groups.
In addition, extended context alone is insufficient because a single global prompt cannot reliably control scene transitions or orchestrate events at precise time points in long videos. We therefore introduce shot-level textual conditioning via cross-modal attention, where each shot is guided by a concise description~\citep{gu2025long,wang2025echoshot}. 
To support this, we build a data pipeline that produces minute-level video samples with dense, multi-shot captions and reliable shot segmentation.

Our contributions: 
We propose MoGA, an effective sparse attention mechanism that replaces block-level scoring with precise group assignment via a lightweight token router, enabling effective modeling of long contexts.
Building on MoGA, we introduce a video generation model capable of producing minute-level, multi-shot, 480p videos at 24 fps with a context length of about 580k tokens. Fig. \ref{1_min_video} illustrates a one-minute video generated by our model. 
Extensive evaluations show consistent improvements over state-of-the-art (SoTA) sparse attention baselines and a multi-shot video generation model.

\section{Related Work}

\subsection{Long Video Generation}
Previous work on long video generation beyond typical duration limits has converged on three main paradigms.
\textbf{Multistage} methods decompose long video generation into multiple steps~\citep{yin2023nuwa,zhuang2024vlogger,tian2024videotetris,huang2025filmaster, xiao2025captain,wang2025storyanchors}.
For example, Captain Cinema~\citep{xiao2025captain} adopts hierarchical planning with top-down keyframe generation and bottom-up synthesis for narrative coherence. 
Multistage approaches introduce hand-crafted inductive biases and pose challenges for end-to-end optimization.
\textbf{Autoregressive} approaches generate videos through sequential segment synthesis~\citep{chen2024diffusion,huang2025selfforcingbridgingtraintest,yin2025slow,henschel2025streamingt2v,gu2025long,ai2025magi1autoregressivevideogeneration}. 
Diffusion Forcing~\citep{chen2024diffusion} adapts denoising schedules for variable sequence lengths.
CasusVid~\citep{yin2025slow} distills bidirectional models into an efficient autoregressive model.
StreamingT2V~\citep{henschel2025streamingt2v} combines short- and long-term memory for streaming video extension. 
FAR~\citep{gu2025long} introduces hierarchical causal representations for multiscale dependencies. 
MAGI-1~\citep{ai2025magi1autoregressivevideogeneration} demonstrates the scaling capability of this paradigm.
\textbf{Context compression} methods address computational constraints by compressing historical content~\citep{dalal2025one,zhang2025packing,jiang2025lovicefficientlongvideo}. TTT~\citep{dalal2025one} compresses long context via a bidirectional recurrent layer. 
FramePack~\citep{zhang2025packing} employs importance-based frame compression to maintain a fixed computational budget. 
However, these methods either produce videos of limited duration~\citep{chen2024diffusion,huang2025selfforcingbridgingtraintest,ai2025magi1autoregressivevideogeneration} or fail to generate multi-shot videos in real-world scenes~\citep{yin2025slow,henschel2025streamingt2v,gu2025long,dalal2025one,zhang2025packing,jiang2025lovicefficientlongvideo}.
A closely related line of work is LCT~\citep{guo2025long}, which models interleaved multi-shot prompts and videos within a local context window using \emph{}{full attention}.
While pioneering end-to-end multi-shot long video generation, LCT remains constrained by the quadratic cost of full attention.

\subsection{Sparse Attention for Video Generation}
Attention–based foundation models unify many domains and consistently exhibit a common sparsity structure~\citep{lu2025moba,yuan2025native,deepseekai2024deepseekv32}.
In video generation, given the inherent sparsity, a natural approach to efficient generation is to select important query-key pairs.
Prior work broadly falls into two categories: static priors~\citep{zhang2025fast,xi2025sparse,li2025radial} and coarse-to-fine dynamic routing~\citep{wu2025vmoba,yang2025sparse,zhang2025faster}. 
Among static approaches, STA~\citep{zhang2025fast} employs 3D sliding windows with a hardware-aware implementation.
SVG~\citep{xi2025sparse} uses online pattern selection to classify attention heads as spatial or temporal sparse attention.
Radial Attention~\citep{li2025radial} introduces a static attention mask to perform spatiotemporal attention with $\mathcal{O}(n \log n)$ complexity.
However, these methods have difficulty modeling evolving long-range dependencies, which are crucial for maintaining cross-shot consistency. 
Another line of work adopts dynamic token routing for sparse attention.
VSA~\citep{zhang2025faster} first obtains compressed representations of contiguous spatiotemporal blocks, and then selects the top‑k blocks for fine-grained attention.
Similarly, VMoBA~\citep{wu2025vmoba} introduces an improved MoBA~\citep{lu2025moba} tailored to video generation.
In such methods, the block size presents a trade-off between expressiveness and efficiency.
Smaller blocks yield more accurate coarse-grained attention estimates but reduce efficiency.
In addition, SVG2~\citep{yang2025sparse} is a training-free dynamic sparse attention method that performs online k-means clustering over tokens during inference and selects the top-k clusters based on their centroids. 
It shares a similar motivation with MoGA, i.e., tokens can be grouped into semantically coherent clusters.
However, online clustering in SVG2 introduces additional k-means computations during the forward pass and is not straightforward to differentiate through.
In contrast, MoGA employs trainable cluster centroids to enable simple and efficient routing with minimal computational overhead, making it suitable for end-to-end training.

\section{Method}

\subsection{Preliminary}

\textbf{Vanilla self-attention} ~\citep{vaswani2017attention} plays a crucial role in video generation with Diffusion Transformers (DiTs) ~\citep{peebles2023scalable}. 
Consider an input sequence $\boldsymbol{X} \in \mathbb{R}^{N \times d}$, where $N = h \times w \times t$ is the total number of tokens across the latent spatial dimensions ($h \times w$) and the latent temporal dimension ($t$), and $d$ denotes the model's hidden dimension. 
For simplicity, we consider a single query case, where $\boldsymbol{x}$ is a token from the input sequence and $\boldsymbol{q}$ is its corresponding query.
Vanilla self-attention (SA) is computed as:
\begin{equation}
\operatorname{SA}(\boldsymbol{q}, \boldsymbol{K}, \boldsymbol{V}) =  \text{softmax}(\frac{\boldsymbol{q}\boldsymbol{K}^{\top}}{\sqrt{d}})\cdot \boldsymbol{V},
\label{eq:3d_attention}
\end{equation}
where $\boldsymbol{K}$ and $\boldsymbol{V}$ denote the keys and values. While self-attention excels at capturing long-range dependencies via global information aggregation, it incurs quadratic computational complexity of $\mathcal{O}(N^2)$. 
The computational burden becomes particularly prohibitive in long-video generation. 
For example, generating a 1-minute video at 480p with approximately 1,600 tokens per frame across 961 frames (16 fps) yields a total token count of about 384k\footnote{Following Wan~\citep{wan2025wan}, the VAE downsampling factors for (t, h, w) are (4, 8, 8) and patchify sizes are (1, 2, 2).}. 
Performing full attention on such a long sequence is intractable.

\begin{figure*}[t]
  \centering
  \includegraphics[width=\linewidth]{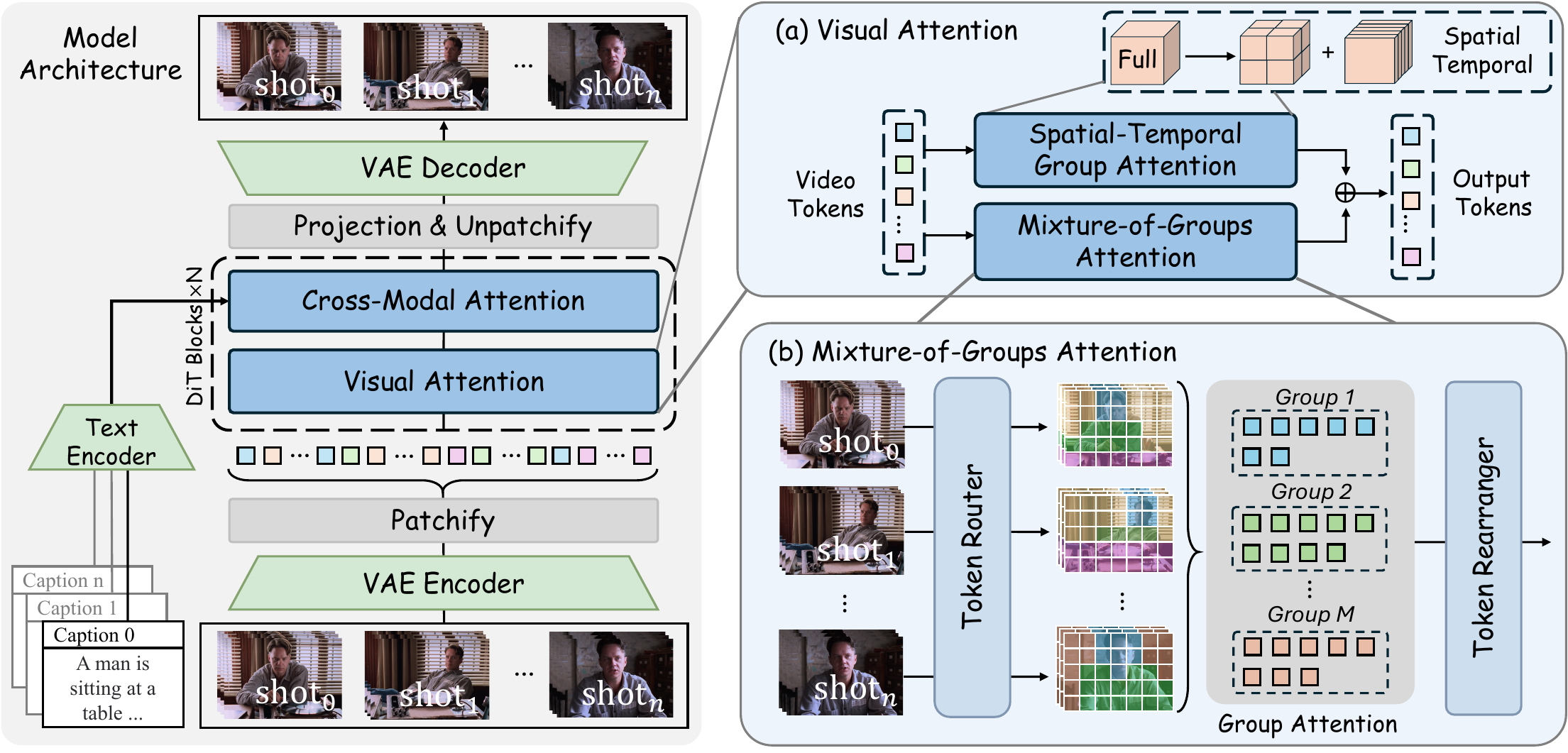}
  \caption{\textbf{Left}: Our model adopts a DiT architecture with interleaved Visual Attention and Cross-Modal Attention blocks. Visual Attention exclusively processes visual content, whereas Cross-Modal Attention enables shot-level text conditioning, instantiated via either cross-attention~\citep{wan2025wan} or multi-modal attention~\citep{kong2024hunyuanvideo,esser2024scaling}. \textbf{Top-right} (a): Visual Attention combining MoGA with Spatial‑Temporal Group Attention for global-local consistency. \textbf{Bottom‑right} (b): MoGA, where a router groups tokens and performs intra-group attention, enabling long-range global interactions.}
  \label{fig:framework}
  \vspace{-5pt}
\end{figure*}

In this section, we introduce MoGA for efficient long video generation. 
The overall architecture is shown in Fig. \ref{fig:framework}. 
We first present the preliminaries, then detail MoGA, and finally describe the pipeline for constructing multi-shot long-video training data.

Beyond computational cost, full attention is not ideally aligned with the structure of videos. 
In videos, softmax attention is inherently sparse~\citep{xi2025sparse} because nearby tokens exhibit strong local spatiotemporal correlation, while only a few globally shared, dynamic semantics persist across frames.
Most query–key pairs contribute little, whereas a small subset dominates~\citep{ge2023model}.
For long videos, attention should leverage this sparsity by prioritizing important query–key interactions to reduce redundancy.

\subsection{Mixture-of-Groups Attention (MoGA)}

\textbf{MoGA} addresses the above challenge via efficient token routing, where a lightweight, trainable router assigns correlated tokens to groups and performs self-attention within each group.
Specifically, the router is a linear projection followed by softmax gating, similar to MoE~\citep{fedus2022switch}.
Given a token $\boldsymbol{x} \in \mathbb{R}^d$ and a predetermined number of groups $M$, the router computes routing scores $\boldsymbol{r} \in \mathbb{R}^M$ as:
\begin{equation}
\boldsymbol{r} = \operatorname{Router}(\boldsymbol{x}).
\end{equation}
The group assignment probabilities are computed as:
\begin{equation}
p(i \mid \boldsymbol{x}) = \operatorname{softmax}(\boldsymbol{r})_i,
\end{equation}
and the token is assigned to the group with the highest probability:
\begin{equation}
g(\boldsymbol{x}) = \arg\max_{i \in [M]} \; p(i  \mid  \boldsymbol{x}).
\end{equation}
Following group assignment, we apply self-attention independently within each group. The MoGA output is:
\begin{equation}
\operatorname{MoGA}(\boldsymbol{x}) = p(g(\boldsymbol{x}) \mid \boldsymbol{x}) \cdot \operatorname{SA}(\boldsymbol{q},\boldsymbol{K}_{g(\boldsymbol{x})},\boldsymbol{V}_{g(\boldsymbol{x})}),
\end{equation}
where $\boldsymbol{K}_{g(\boldsymbol{x})}$ and $\boldsymbol{V}_{g(\boldsymbol{x})}$ are the keys and values of the group $g(\boldsymbol{x})$, and $\boldsymbol{q}$ is the query feature of $\boldsymbol{x}$.
This grouped attention mechanism reduces computational complexity from $\mathcal{O}(N^2)$ to a theoretical minimum of $\mathcal{O}(N^2/M)$ under uniform group assignment.

\begin{figure*}[t]
  \centering
  \includegraphics[width=0.9\linewidth]{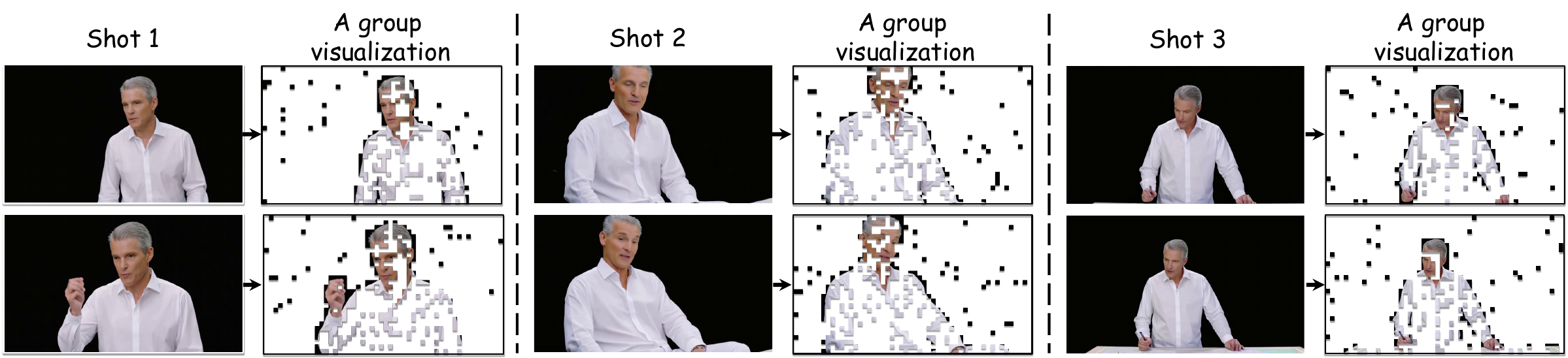}
  \vspace{-5pt}
  \caption{Visualization of dynamic router grouping.}
  \label{group_ablation}
\end{figure*}

As illustrated in Fig. \ref{group_ablation}, we extract the grouping assignments from an intermediate-layer router during the video generation process and visualize one representative group. After end-to-end training, the router assigns the man’s head, hands, and portions of his clothing to the same group, indicating its ability to capture semantically coherent structures that span shot boundaries. 

MoGA builds on groupwise attention and remains compatible with high-performance kernels such as FlashAttention~\citep{dao2023flashattention} (see Alg.~\ref{alg:1}). Beyond sparse attention, a second pillar of long-context modeling is sequence parallelism~\citep{jacobs2023deepspeed}, with which MoGA is also compatible. 
Before the sequence gather and head scatter step in each attention layer, MoGA computes routing scores over tokens (with whole heads) and then aggregates the routing results across all tokens.

\begin{algorithm}[t]
\caption{MoGA Pseudocode with FlashAttention}
\begin{algorithmic}[1]
    \Require \(\boldsymbol{Q},\boldsymbol{K},\boldsymbol{V}\) are the query, key and value of tokens \(\boldsymbol{X}\) 
    \State \(\boldsymbol{g} = \text{router}(\boldsymbol{X})\) \Comment{MoGA routing results}
    \State \(\hat{\boldsymbol{Q}}, \hat{\boldsymbol{K}}, \hat{\boldsymbol{V}}, \text{cu\_seqlen}, \text{max\_seqlen}, \text{permute\_index} = \text{permute}(\boldsymbol{Q}, \boldsymbol{K}, \boldsymbol{V}, \boldsymbol{g})\)
    \State \(\hat{\boldsymbol{O}} = \text{flash\_attn} (\hat{\boldsymbol{Q}}, \hat{\boldsymbol{K}}, \hat{\boldsymbol{V}}, \text{cu\_seqlen}, \text{max\_seqlen})\)
    \State \(\boldsymbol{O}= \text{repermute} (\hat{\boldsymbol{O}},\text{permute\_index}) \) \Comment{MoGA recovers the original token positions}
\end{algorithmic}
\label{alg:1}
\end{algorithm}

\textbf{Group Balancing Loss.} A potential issue with token assignment is that the router may collapse by routing most tokens to only a few groups, which would degenerate MoGA into full attention
To encourage adaptive token allocation across groups, we introduce an auxiliary \textit{group balancing loss}, inspired by the load balancing loss~\citep{fedus2022switch} used in MoE. The loss is defined as:
\begin{equation}
\label{bloss}
\mathcal{L}_{\text{gb}} = \alpha \cdot M  \sum_{i=1}^{M} F_i  P_i,
\end{equation}
where $\alpha$ is a loss weight and $F_i$ is the fraction of tokens allocated to group $i$, 
\begin{equation}
F_i = \frac{1}{N} \sum_{\boldsymbol{x}} \mathbf{1}(g(\boldsymbol{x})=i),
\end{equation}
where $\mathbf{1}$ is the indicator function, and $P_i$ is the mean routing probability allocated for group $i$,
\begin{equation}
P_i = \frac{1}{N} \sum_{g(\boldsymbol{x})=i}  p(g(\boldsymbol{x}) \mid \boldsymbol{x}).
\end{equation}
Minimizing $\mathcal{L}_{\text{gb}}$ encourages uniform token assignment across groups, as this objective attains its minimum under a uniform distribution~\citep{fedus2022switch}.

\textbf{Spatial-Temporal Group Attention.} 
Although MoGA captures long-range coherence, it lacks local continuity. 
We complement it with local spatiotemporal group attention (STGA)~\citep{gao2025seedance, zhang2025waver}, which restricts self‑attention to local windows in latent video space, as shown in Fig. \ref{fig:framework}(a). 
This captures short‑range dependencies with bounded compute.

We first partition the latent video into fixed spatial windows and then group frames along the temporal axis. 
Frames from different shots are assigned to distinct temporal groups.
We empirically find that completely removing inter‑shot interactions causes flicker in the first frame after a shot cut. 
To mitigate this, when computing group attention, we augment the keys and values with two latent frames from adjacent shots (without augmenting the queries).
This preserves continuity at shot boundaries with negligible additional compute.
To enable intra-frame information exchange, we also perform per-frame attention by grouping tokens within each latent frame.
Each token therefore receives outputs from multiple groups (one dynamic and two static), and we take their mean as the final output. 

\begin{figure*}[t]
  \centering
  \includegraphics[width=\linewidth]{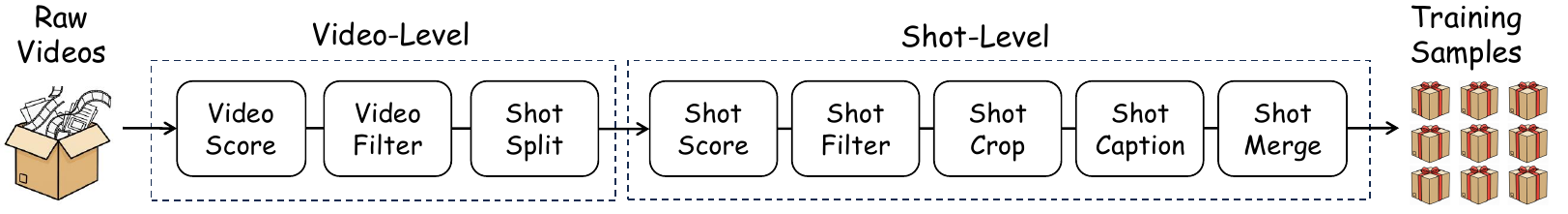}
  \vspace{-15pt}
  \caption{Multi-shot long video data pipeline.}
  \label{datapipe}
  \vspace{-10pt}
\end{figure*}

\subsection{Data Pipeline}
\label{data}

We construct a pipeline that converts raw long videos into one‑minute, multi‑shot clips with dense annotations for long video generation.
The pipeline has two stages: a video-level phase and a shot-level phase (Fig. \ref{datapipe}).

\textbf{Video-Level.} We first analyze raw videos using visual quality assessment (VQA) models (e.g., aesthetics~\citep{schuhmann2022laion5b}, clarity, exposure) and simple operators (e.g., black-border detection) to obtain metadata and quality scores.
We then filter raw videos with source‑specific, calibrated thresholds to remove low‑quality content. 
Because long video samples require temporal coherence, we relax clip-level filtering~\citep{zheng2024open, kong2024hunyuanvideo} while applying stricter filtering at the source (raw-video) level.
Next, we segment each video into single-shot clips using AutoShot~\citep{zhu2023autoshot} and PySceneDetect~\citep{PySceneDetect}.
AutoShot shows higher sensitivity to fades and gradual transitions.
Combining predictions from both tools allows us to label whether a boundary is clean or affected by transition overlap.
This stage yields a pool of single‑shot clips.

\textbf{Shot-Level.} 
We process single-shot clips using VQA and optical character recognition (OCR) models and discard low-quality clips.
Based on OCR results, we compute a maximum-area crop that excludes watermarks and subtitles while preserving the original aspect ratio.
Clips with insufficient retained area are discarded.
Next, we generate captions for cropped clips using a multimodal large language model~\citep{bai2025qwen2}. 
Finally, we merge temporally adjacent single-shot clips into multi-shot training samples (up to 65 seconds) and trim a few frames from clips affected by transition overlap to ensure clean boundaries.

\section{Experiments}

\textbf{Training Settings.} 
We fine-tune MoGA on existing DiT-based short video generation models with the rectified flow objective~\citep{esser2024scaling}.
For a fair comparison with baselines, we train MoGA on the open-source Wan2.1 models (1.3B and 14B)~\citep{wan2025wan}.
The resulting model stably generates 477 frames at 16 fps (30 seconds) and 480p resolution, with a context length of 187k. 
We use a constant learning rate of 1$e$-5.
The loss weight $\alpha$ is set to $0.1$. 
We set the number of groups to $M = 5$ and partition the spatial grid into $2\times2$ groups.
We adopt a multistage training strategy: 3k steps on 10-second clips followed by 1k steps on 30-second clips. 

Because MoGA is a general sparse attention, we also apply it to a video generation model built on MMDiT~\citep{esser2024scaling,kong2024hunyuanvideo,gao2025seedance}. 
Unlike Wan, this model replaces cross-attention with MMDiT to perform cross-modal attention. 
It partitions space into $4\times4$ groups and sets the router's group number to $M = 20$, enabling a much longer context length.
The MMDiT-based model generates 1,441 frames at 24 fps (60 seconds) at 480p, with a context length of 578k. 

\textbf{Baselines.}  
To evaluate our method, we compare with multiple baselines. 
For multi-shot long video generation, we include the keyframe-based pipeline IC-LoRA+Wan~\citep{huang2024context, wan2025wan} and EchoShot~\citep{wang2025echoshot}, which natively supports multi-shot generation. 
For sparse video generation, we compare against sparse attention methods, including the training-based DiTFastAttn \citep{yuan2024ditfastattn} and the training-free methods SVG \citep{xi2025sparse} and VMoBA~\citep{wu2025vmoba}.

\textbf{Evaluation Metrics.}
Following prior work, we evaluate all methods using the metrics introduced by VBench~\citep{huang2024vbench}. 
Specifically, subject consistency and background consistency measure how well the main subjects and backgrounds of sampled frames are preserved throughout the video. 
Motion smoothness measures motion fluidity, penalizing jitter and abrupt transitions.
We also report aesthetic quality and image quality to quantify the visual appeal and technical fidelity of each frame.
To compute cross-shot consistency, we first sample a fixed number of frames from different shots.
We then compute feature similarities across shots using CLIP~\citep{radford2021learning} and DINOv2~\citep{oquab2023dinov2}, referred to as Cross‑Shot CLIP and Cross‑Shot DINO.
For single-shot 5-second video generation, we constructed a diverse test set comprising 300 prompts.
For multi-shot 10-second video generation, we use the 100 multi-shot prompt sets from~\citep{wang2025echoshot}. 
For long video generation, we evaluate on a test set of 11 scripts comprising 105 prompts. 
Each script contains 8–10 shots to produce a 30-second video.


\subsection{Quantitative Results}

\begin{table}[t]
\centering
\setlength{\tabcolsep}{8pt} 
\renewcommand{\arraystretch}{1} 
\adjustbox{width=\textwidth}{%
\begin{tabular}{lccccccc}
\toprule
\multicolumn{1}{c}{\textbf{Method}} &
\multicolumn{1}{c}{\textbf{Base Model}} &
\multicolumn{1}{c}{\begin{tabular}[c]{@{}c@{}}\textbf{Subject}\\ \textbf{Consistency} $\uparrow$\end{tabular}} &
\multicolumn{1}{c}{\begin{tabular}[c]{@{}c@{}}\textbf{Background}\\ \textbf{Consistency} $\uparrow$\end{tabular}} &
\multicolumn{1}{c}{\begin{tabular}[c]{@{}c@{}}\textbf{Motion}\\ \textbf{Smoothness} $\uparrow$\end{tabular}} &
\multicolumn{1}{c}{\begin{tabular}[c]{@{}c@{}}\textbf{Aesthetic}\\ \textbf{Quality} $\uparrow$\end{tabular}} &
\multicolumn{1}{c}{\begin{tabular}[c]{@{}c@{}}\textbf{Image}\\ \textbf{Quality} $\uparrow$\end{tabular}} &
\multicolumn{1}{c}{\textbf{Sparsity} $\uparrow$} \\
\midrule
Wan (Original) & Wan2.1-14B & 0.9611 & \textbf{0.9560} & \textbf{0.9936} & 0.5807 & 0.6680 & 0\%  \\
\midrule
DiTFastAttn (Training-based) & Wan2.1-14B &0.9456 & 0.9394 & 0.9924 & 0.5269 & 0.6466 & 50.00\% \\
SVG (Training-free) & Wan2.1-14B & 0.9002  & 0.8926  & 0.9870  & 0.5370  &  0.6357 & 50.00\%  \\
VMoBA (Training-free)  & Wan2.1-14B   &  0.8605 & 0.8876 & 0.9789 & 0.5369 & 0.6111 & 31.00\% \\
\midrule
MoGA (Ours) & Wan2.1-14B & \textbf{0.9699}  & 0.9542 & 0.9927 & \textbf{0.5810} & \textbf{0.6994} & \textbf{71.25\%} \\
\bottomrule
\end{tabular}%
}
\vspace{-5pt}
\caption{Quantitative comparison for 5-second single-shot video generation.}
\label{tab:metrics2}
\end{table}

\begin{table}[t]
\centering
\label{tab:metrics}
\setlength{\tabcolsep}{4pt} 
\renewcommand{\arraystretch}{1} 
\adjustbox{width=0.95\textwidth}{%
\begin{tabular}{lcccccccc}
\toprule
\multicolumn{1}{c}{\textbf{Method}} &
\multicolumn{1}{c}{\textbf{Base Model}} &
\multicolumn{1}{c}{\begin{tabular}[c]{@{}c@{}}\textbf{Subject}\\ \textbf{Consistency} $\uparrow$\end{tabular}} &
\multicolumn{1}{c}{\begin{tabular}[c]{@{}c@{}}\textbf{Background}\\ \textbf{Consistency} $\uparrow$\end{tabular}} &
\multicolumn{1}{c}{\begin{tabular}[c]{@{}c@{}}\textbf{Motion}\\ \textbf{Smoothness} $\uparrow$\end{tabular}} &
\multicolumn{1}{c}{\begin{tabular}[c]{@{}c@{}}\textbf{Aesthetic}\\ \textbf{Quality} $\uparrow$\end{tabular}} &
\multicolumn{1}{c}{\begin{tabular}[c]{@{}c@{}}\textbf{Image}\\ \textbf{Quality} $\uparrow$\end{tabular}}  &
\multicolumn{1}{c}{\begin{tabular}[c]{@{}c@{}}\textbf{Cross-Shot}\\ \textbf{DINO} $\uparrow$\end{tabular}}  &
\multicolumn{1}{c}{\begin{tabular}[c]{@{}c@{}}\textbf{Cross-Shot}\\ \textbf{CLIP} $\uparrow$\end{tabular}}  \\
\midrule
IC-Lora+Wan & Wan2.1-1.3B & 0.9476 & 0.9538  &  0.9901 & 0.5237  &  0.6684 & 0.4669 & 0.7169  \\
EchoShot & Wan2.1-1.3B  & 0.9544 & 0.9518 &  \textbf{0.9939} &  0.5718 & 0.6534  & 0.5961 & 0.8469   \\
\midrule
MoGA (Ours) & Wan2.1-1.3B & \textbf{0.9549} & \textbf{0.9597} & 0.9919 & \textbf{0.5890} & \textbf{0.6729} & \textbf{0.6623}  &  \textbf{0.8654}   \\
\bottomrule
\end{tabular}%
}
\vspace{-5pt}
\caption{Quantitative comparison for 10-second multi-shot video generation.}
\label{long_video_score_10}
\end{table}

\begin{table}[!tbp]
\centering
\setlength{\tabcolsep}{4pt} 
\renewcommand{\arraystretch}{1} 
\adjustbox{width=0.8\textwidth}{%
\begin{tabular}{lcccccc}
\toprule
\multicolumn{1}{c}{\textbf{Method}} &
\multicolumn{1}{c}{\textbf{Base Model}} &
\multicolumn{1}{c}{\begin{tabular}[c]{@{}c@{}}\textbf{Subject}\\ \textbf{Consistency} $\uparrow$\end{tabular}} &
\multicolumn{1}{c}{\begin{tabular}[c]{@{}c@{}}\textbf{Background}\\ \textbf{Consistency} $\uparrow$\end{tabular}} &
\multicolumn{1}{c}{\begin{tabular}[c]{@{}c@{}}\textbf{Motion}\\ \textbf{Smoothness} $\uparrow$\end{tabular}} &
\multicolumn{1}{c}{\begin{tabular}[c]{@{}c@{}}\textbf{Aesthetic}\\ \textbf{Quality} $\uparrow$\end{tabular}} &
\multicolumn{1}{c}{\begin{tabular}[c]{@{}c@{}}\textbf{Image}\\ \textbf{Quality} $\uparrow$\end{tabular}} \\

\midrule
IC-Lora+Wan & Wan2.1-14B & 0.8946 & 0.9169  & 0.9872 & 0.5759  & 0.6835   \\
\midrule
MoGA (Ours) & Wan2.1-14B & \textbf{0.9572} & \textbf{0.9475} & 0.9893 & 0.5789 & 0.6993   \\
MoGA (Ours) & MMDiT & 0.9305 & 0.9301 & \textbf{0.9895} & \textbf{0.5881} & \textbf{0.6996}    \\
\bottomrule
\end{tabular}%
}
\vspace{-5pt}
\caption{Quantitative comparison for 30-second multi-shot long video generation.}
\label{long_video_score}
\vspace{-10pt}
\end{table}

\begin{figure*}[t]
  \centering
  \includegraphics[width=0.8\linewidth]{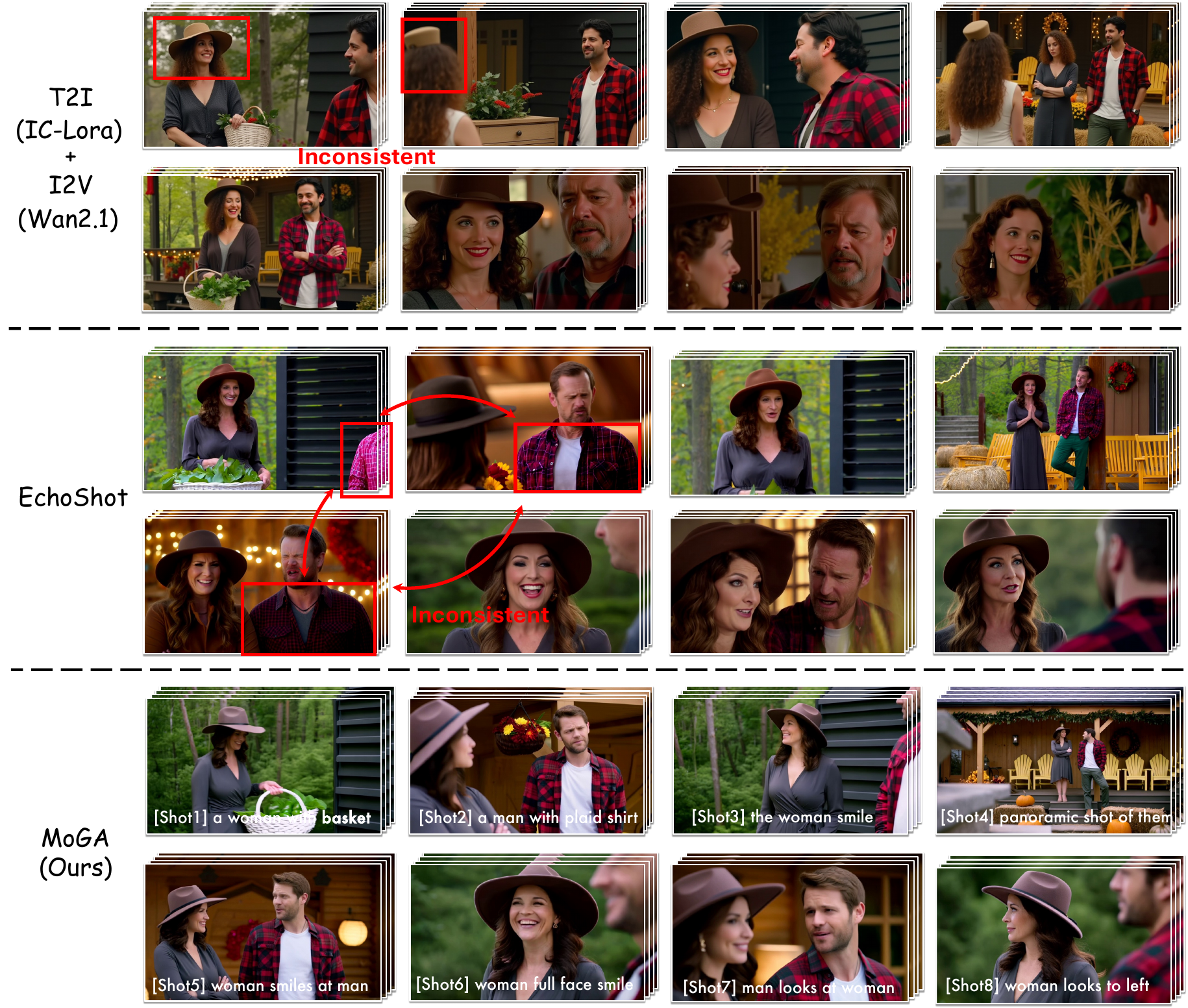}
  \vspace{-5pt}
  \caption{Qualitative results of MoGA and other methods. We present eight representative shots, demonstrating long-content coherence, character consistency, and visual quality.}
  \label{fig:showcase}
\end{figure*}
\begin{figure*}[!tbp]
  \centering
  \includegraphics[width=0.8\linewidth]{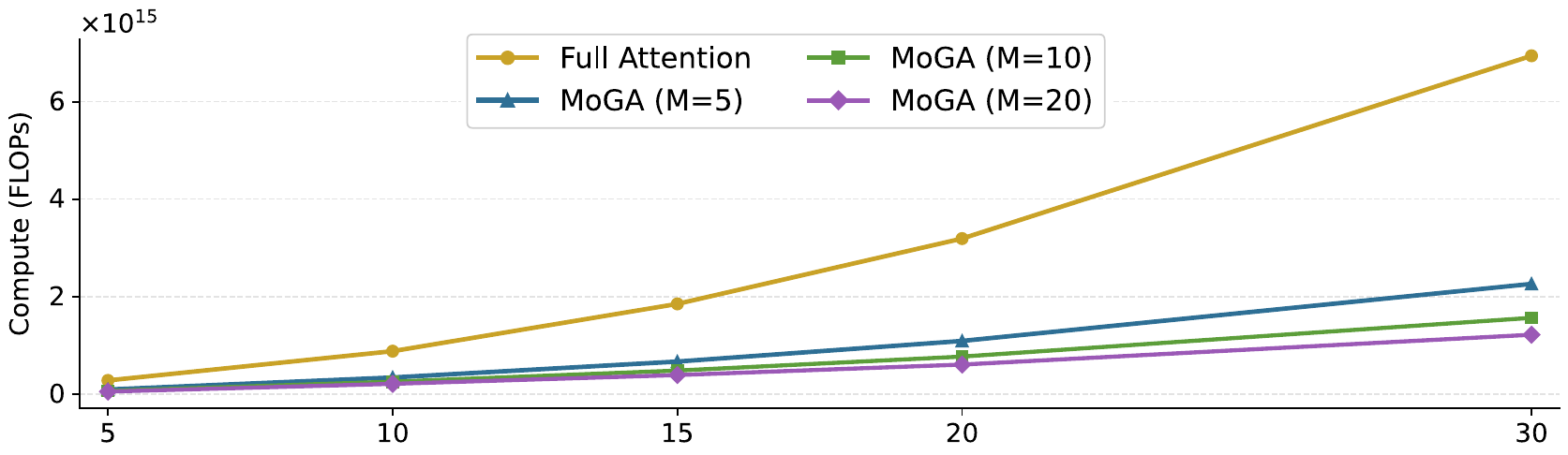}
  \vspace{-5pt}
  \caption{Computational efficiency. The x-axis denotes the generated video duration (s). As the number of groups~($M$) increases, MoGA's FLOPs decrease substantially. }
  \label{GFLOPs_curve}
\end{figure*}

First, we compare MoGA with prior sparse attention methods for single-shot, short video generation, following their evaluation settings to ensure fairness. As shown in Tab.~\ref{tab:metrics2}, despite higher sparsity, MoGA achieves consistent improvements over existing sparse baselines across metrics. 
It is worth noting that although our method is highly sparse, it can still match or surpass the original Wan (full attention) on multiple metrics.

Next, we compare MoGA with other multi-shot video generation methods.
Tab.~\ref{long_video_score_10} reports quantitative comparisons among MoGA, IC-LoRA+Wan and EchoShot. 
Despite relying on sparse attention, our method outperforms the full attention baseline (EchoShot) on most metrics, indicating that preserving interactions among salient tokens not only reduces FLOPs but also suppresses noise from irrelevant content. 
This leads to stronger character identity consistency and improved temporal scene coherence.

Finally, we benchmark long video generation against the baseline. 
Because few open-source methods can produce 30-second, multi-shot videos, we compare MoGA (with two backbones) to IC-LoRA+Wan. As shown in Tab.~\ref{long_video_score}, MoGA substantially outperforms IC-LoRA+Wan under the same backbone, highlighting the benefits of end-to-end modeling over multistage pipelines. Notably, even under aggressive sparsity, MoGA with MMDiT maintains high visual fidelity, indicating a scalable path to longer context lengths.

\subsection{Qualitative Results}

\textbf{Visual Comparison.} In this subsection, we present qualitative results on 30-second videos across representative baselines.
Because EchoShot cannot natively produce 30-second outputs, we concatenate video clips generated by EchoShot to form the full sequence.
As shown in Fig. \ref{fig:showcase}, the IC-LoRA+Wan pipeline is constrained by its per-iteration image cap (typically three frames), which limits its ability to cover a larger number of shots. Consequently, it often exhibits subject drift and background inconsistency as the sequence progresses. 
EchoShot scales to more shots but still shows notable cross-shot inconsistencies on long temporal ranges. In contrast, MoGA maintains stable, coherent content over extended durations. 
For example, even without repeated or explicit specification across shots, the woman’s hat remains consistently preserved. Since STGA lacks explicit cross-shot information exchange, this consistency can be attributed to MoGA, which effectively selects and maintains shot-spanning identity and context. 

\begin{figure*}[t]
  \centering
  \includegraphics[width=0.95\linewidth]{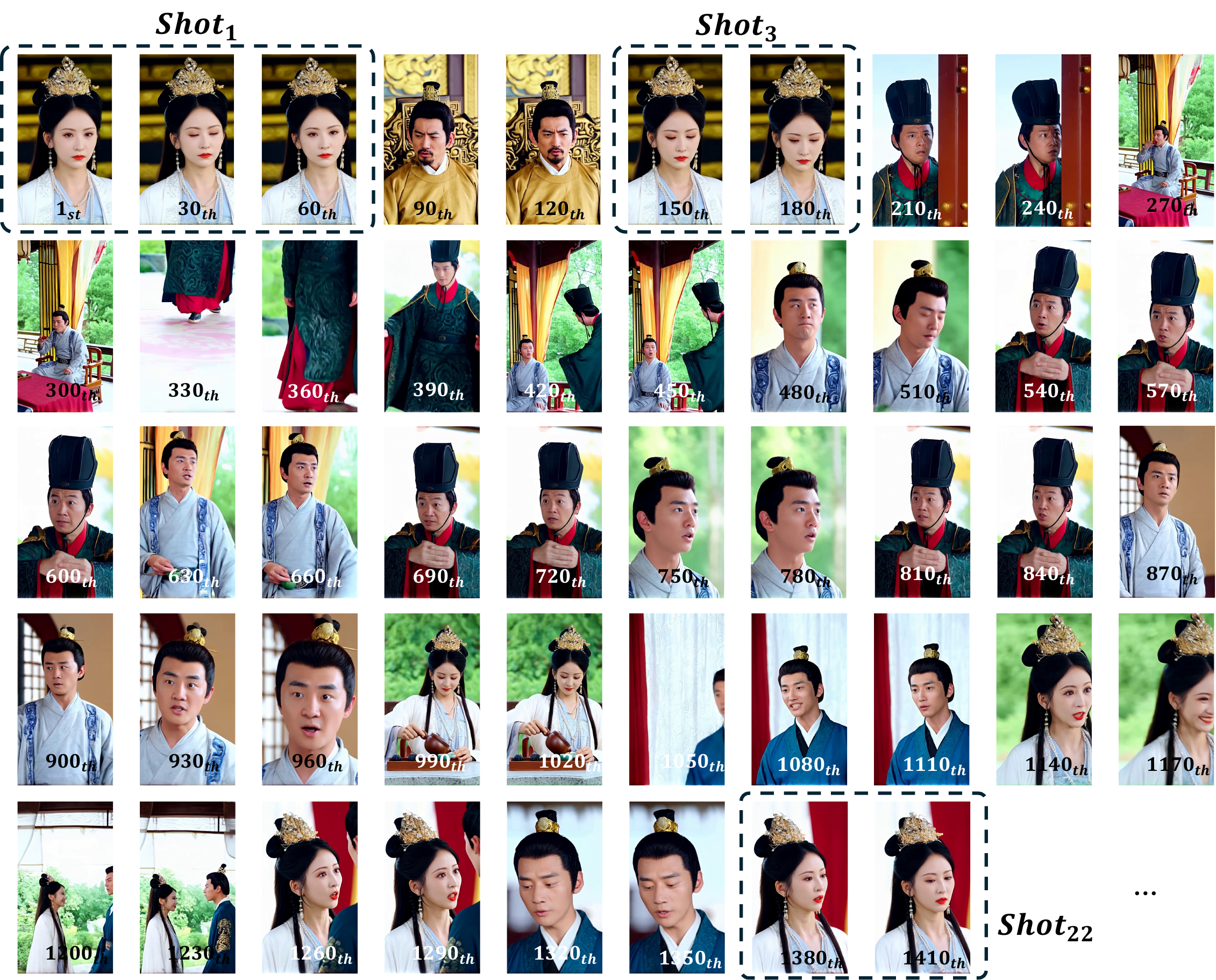}
  \caption{One-minute video generated by MoGA.}
  \label{1_min_video}
  \vspace{-2mm}
\end{figure*}

\textbf{One-Minute Video of 1,441 Frames.} In Fig. \ref{1_min_video}, we present the generated results of MoGA on an ultra-long video of over one minute, using the MMDiT-based MoGA model ($M = 20$). 
MoGA maintains strong long-range contextual consistency. 
The 1st and 22nd shots remain highly coherent, and fine details such as the woman’s hairpin and earrings are preserved across shots. Moreover, even with multiple faces appearing across different shots, the model avoids identity confusion. 

\begin{figure*}[h]
  \centering
  \includegraphics[width=0.95\linewidth]{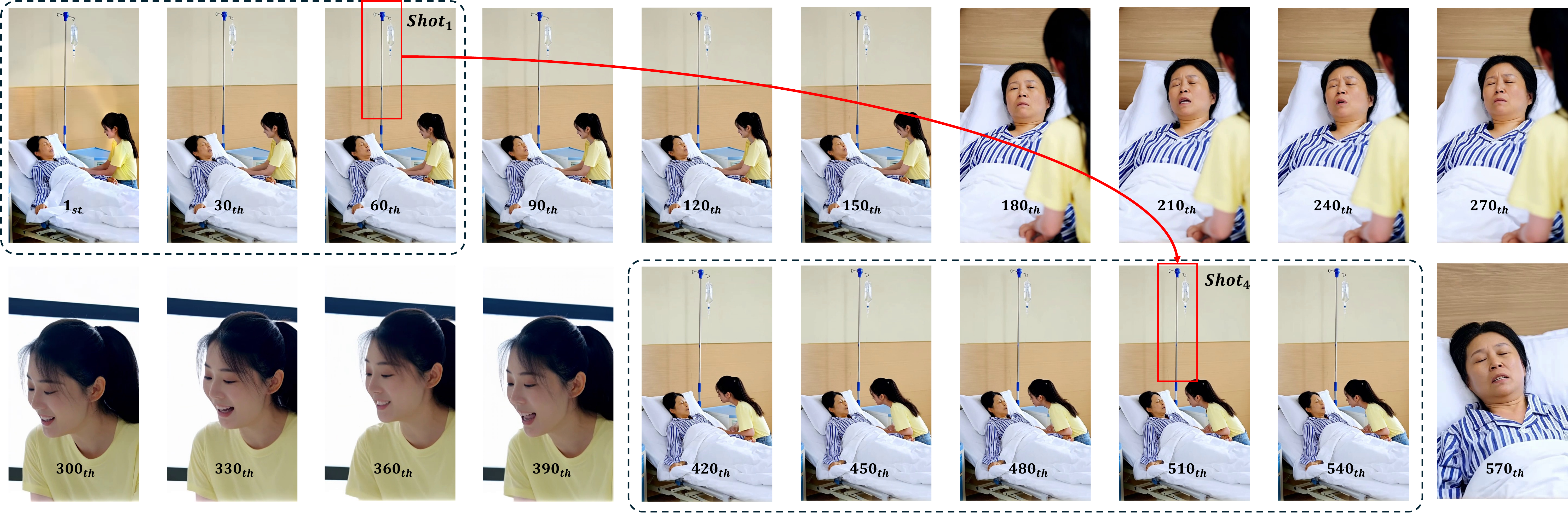}
  \caption{Emergence of background consistency.}
  \label{fig:Background_consistency}
\end{figure*}

\textbf{Emergence of Background Consistency.}
As shown in Fig. \ref{fig:Background_consistency}, we demonstrate MoGA's ability to maintain background consistency. 
After training on long, multi-shot videos, MoGA exhibits emergent, implicit control over consistency in both the environment and the characters. 
Even without explicit specification of details (e.g., the cabinet shape and the position of intravenous drip bottle), different shots automatically maintain coherent, temporally consistent depictions.

\begin{figure*}[h]
  \centering
  \includegraphics[width=0.95\linewidth]{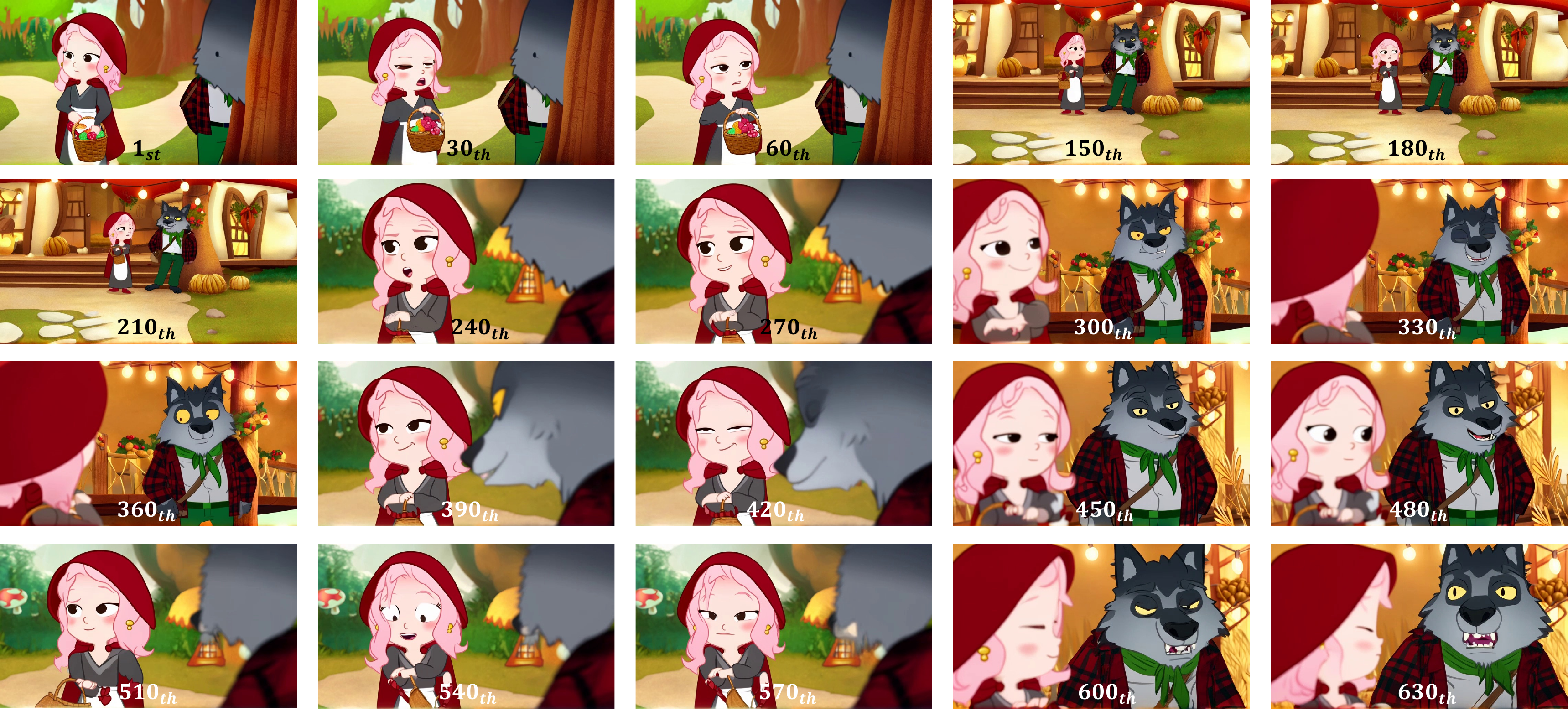}
  \caption{Animation style generation for MoGA.}
  \label{fig:multistyle}
  \vspace{-2mm}
\end{figure*}

\textbf{Multi-Style Video Generation.}
Fig. \ref{fig:multistyle} illustrates MoGA’s multi‑style generation capability. 
MoGA not only performs strongly in realistic spaces but also excels in stylized domains such as animation. 
It can produce high‑quality, long-form 2D videos while maintaining temporal coherence, identity consistency, and scene continuity across diverse styles.

\subsection{Ablation Study}

\textbf{Computational Efficiency.} Fig.~\ref{GFLOPs_curve} plots the relationship between the number of groups ($M$) and FLOPs for the Wan2.1-1.3B model. Our experiments show that even with a relatively small group count ($M{=}5$) for 30-second videos, MoGA achieves substantial computational savings compared to full attention (2.26 PFLOPs vs. 6.94 PFLOPs). It also delivers a $1.7\times$ speedup in both training and inference. Notably, unlike alternative sparse attention such as VMoBA, which incur additional memory overhead due to their block-based mechanisms, our approach maintains memory efficiency without additional memory consumption. 

\textbf{Routing Group Number $M$.} We conduct an ablation study on the number of groups under a fixed computational budget (Tab.~\ref{ablation_of_group_num}). 
Cross-shot DINO and CLIP scores exhibit a rise-then-fall trend as the number of groups increases. This suggests that a moderate level of grouped sparsity strikes a balance between global consistency and efficiency, yielding near‑optimal consistency while maintaining computational efficiency.

\begin{table}[t]
\centering
\setlength{\tabcolsep}{10pt}
\renewcommand{\arraystretch}{1}
\adjustbox{width=0.75\textwidth}{
\begin{tabular}{cccccc}
\toprule
\textbf{Group Numbers} & \textbf{Cross-Shot DINO} $\uparrow$ & \textbf{Cross-Shot CLIP} $\uparrow$  & \textbf{Sparsity}& \textbf{PFLOPs}\\
\midrule
1  & 0.8206 & 0.5919 & 0\%  & $0.88$ \\
2  & 0.8589 & 0.6761 & 41.25\%  & $0.59$ \\
4  & \textbf{0.8672} & 0.6853 & 66.25\%  & $0.42$ \\
8  & 0.8606 & \textbf{0.6910} & 78.75\%  & $0.36$ \\
16 & 0.8569 & 0.6896 & 81.25\%  &  $0.35$ \\
\bottomrule
\end{tabular}
}
\vspace{-5pt}
\caption{Results of consistency for MoGA with Wan2.1-1.3B on 10-second videos.}
\label{ablation_of_group_num}
\end{table}
\begin{figure*}[t]
  \centering  \includegraphics[width=0.9\linewidth]{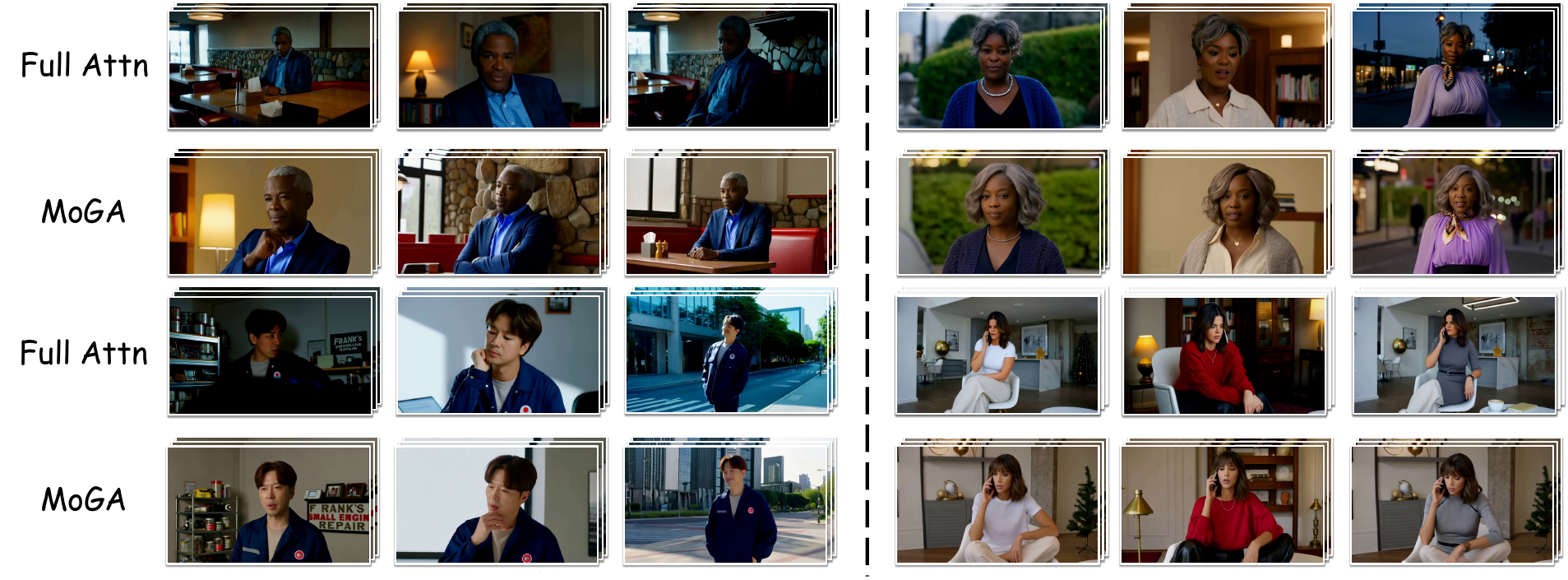}
  \vspace{-5pt}
  \caption{Visual comparison of MoGA vs. full attention for multi-shot generation with a single subject. The left column shows the subject wearing the same outfit across different shots, while the right column shows the subject changing outfits at shot transitions according to the text instructions.}
  \label{moga_full}
\end{figure*}
\begin{figure*}[!tbp]
  \centering
  \includegraphics[width=0.6\linewidth]{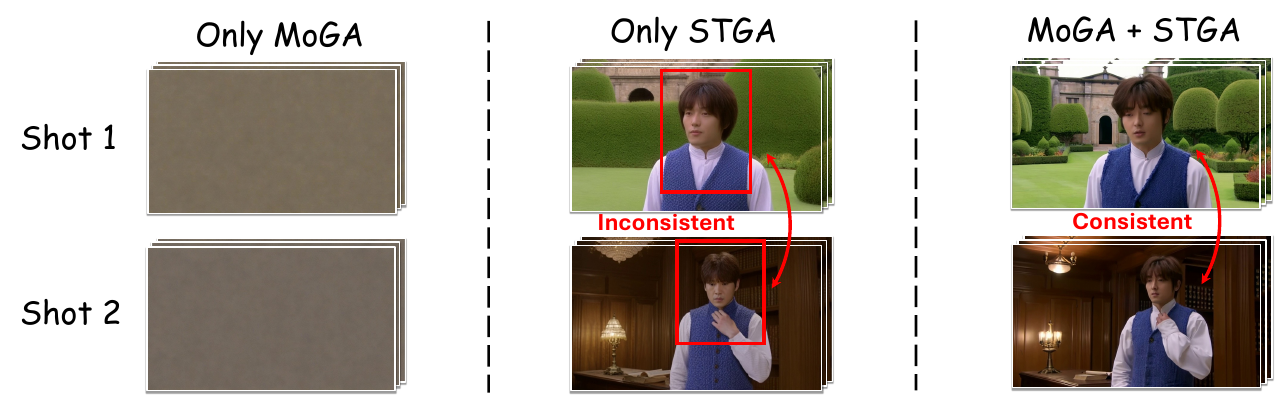}
  \vspace{-5pt}
  \caption{Visual ablation of MoGA and STGA.}
\label{LCGA_STGA_ablation}
  \vspace{-8pt}
\end{figure*}

\textbf{Controllability of Subject Consistency.}
Fig. \ref{moga_full} visualizes a comparison between MoGA and full attention.
Both models are trained on 10‑second data with Wan2.1‑14B. 
The left panel illustrates MoGA’s ability to maintain subject identity across multiple scenes, while the right panel demonstrates its robustness to appearance changes (e.g., clothing) when preserving identity consistency. 
Despite 71.25\% sparsity, MoGA achieves narrative coherence and content editability on par with full attention, and in some cases delivers superior performance.

\textbf{Effectiveness of MoGA and STGA.} As shown in Fig.~\ref{LCGA_STGA_ablation}, MoGA and STGA play complementary roles in enabling context-consistent long video generation. 
Using MoGA alone lacks local information exchange and fails to produce meaningful visual content. Conversely, using only STGA limits long‑range shot interactions, leading to poor cross‑shot consistency and weakened narrative coherence. When combined, the model achieves strong cross‑shot consistency. These results indicate that MoGA effectively routes and preserves shot‑spanning identity and context at relatively low computational cost.

\section{Conclusion}
This paper introduces MoGA, a sparse attention mechanism that replaces coarse block-level scoring with precise, learned group assignments via a lightweight token router. By routing tokens into coherent groups, MoGA improves attention efficiency and fidelity for very long contexts.
Building on MoGA, we propose the video generation model that produces minute-level, multi-shot videos at 480p resolution and 24 fps.
Diverse experiments in video generation further demonstrate the effectiveness of our approach.

\textbf{Acknowledgments.} We thank the ByteDance Seedance team and Wenfeng Lin for their support.

\clearpage

\bibliographystyle{plainnat}
\bibliography{main}

\clearpage

\beginappendix

\section{Details of the Computational Complexity}
As shown in Tab. \ref{tab:pflops_vs_time}, it reports the computational cost under varying number of groups (M) and video duration. As the generation video duration increases, the computational complexity of STGA exhibits approximately linear growth and the computational complexity of MoGA is approximately $1/M$ of that of Full Attention.

\begin{table}[h]
\centering
\setlength{\tabcolsep}{8pt}
\renewcommand{\arraystretch}{1.2}
\begin{tabular}{llccccc}
\toprule
& Video Duration & 5s & 10s & 15s & 20s & 30s \\
\midrule
& Frames & 77 & 157 & 237 & 317 & 477 \\
& Sequence Length & 31200 & 62400 & 93600 & 124800 & 187200 \\
\midrule
\multirow{4}{*}{PFLOPs} & Full Attention & 0.28 & 0.88 & 1.85 & 3.19 & 6.94 \\
& MoGA ($M=5$) & 0.093 & 0.34 & 0.67 & 1.09 & 2.26 \\
& MoGA ($M=10$) & 0.065 & 0.25 & 0.48 & 0.78 & 1.56 \\
& MoGA ($M=20$) & 0.051 & 0.21 & 0.39 & 0.61 & 1.22 \\
\bottomrule
\end{tabular}
\caption{Compute (PFLOPs) versus group number $M$ and video duration on Wan2.1-1.3B.}
\label{tab:pflops_vs_time}
\end{table}

\section{Analysis of Group Balancing Loss}

\begin{figure*}[h]
  \centering
  \includegraphics[width=0.8\linewidth]{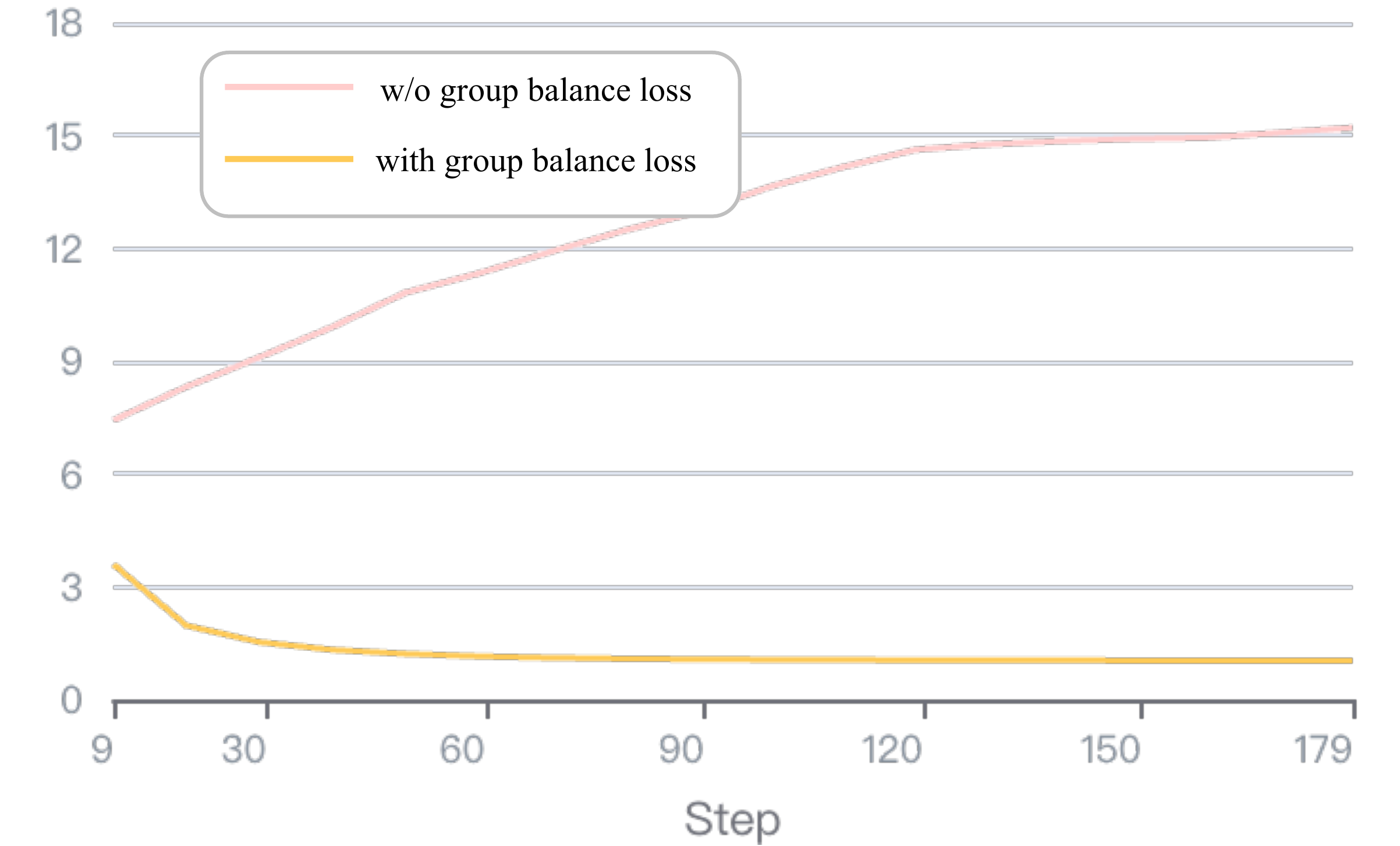}
  \caption{Group balancing loss curves of MoGA.}
  \label{fig:balanceloss}
\end{figure*}

As shown in Fig. \ref{fig:balanceloss}, we validate the effectiveness of the group balancing loss, which measures the balance of the router’s token-to-group assignments. 
A higher value indicates that tokens concentrate in a few groups, whereas a lower value indicates more balanced grouping. 
When we include this loss during training, the metric rapidly converges to around 1, reflecting globally balanced assignments. 
In contrast, without it, the metric increases as the router funnels tokens into a few groups to gain short-term advantages in the diffusion MSE loss. 
Because our goal is to separate weakly related tokens and maintain balanced grouping, the additional group balance loss is necessary to enforce the desired assignments.

\end{document}